\pdfoutput=1

\documentclass[11pt]{article}

\usepackage[]{acl}

\usepackage{times}
\usepackage{latexsym}
\usepackage{multirow}
\usepackage{adjustbox}
\usepackage{amssymb}
\usepackage{enumitem}
\usepackage{graphicx}
\usepackage{amsmath}
\usepackage{amsthm}
\usepackage{booktabs}
\usepackage{bm}
\usepackage{subcaption}

\usepackage[T1]{fontenc}

\usepackage[utf8]{inputenc}

\usepackage{microtype}

\newcommand\blfootnote[1]{%
\begingroup 
\renewcommand\thefootnote{}\footnote{#1}%
\addtocounter{footnote}{-1}%
\endgroup 
}

\title{Knowledge-enhanced Mixed-initiative Dialogue System for Emotional Support Conversations}

\author{
    Yang Deng$^{1}$, Wenxuan Zhang$^{2,\dagger}$, Yifei Yuan$^{1}$, Wai Lam$^{1}$\\
    $^{1}$ The Chinese University of Hong Kong,  $^{2}$ DAMO Academy, Alibaba Group\\ 
    \texttt{
    \{dengyang17dydy,isakzhang\}@gmail.com}\\ \texttt{\{yfyuan,wlam\}@se.cuhk.edu.hk}
}

\begin{document}
\maketitle
\begin{abstract}
Unlike empathetic dialogues, the system in emotional support conversations (ESC)  is expected to not only convey empathy for comforting the help-seeker, but also proactively assist in exploring and addressing their problems during the conversation. 
In this work, we study the problem of mixed-initiative ESC where the user and system can both take the initiative in leading the conversation. 
Specifically, we conduct a novel analysis on mixed-initiative ESC systems with a tailor-designed schema that divides utterances into different types with speaker roles and initiative types. Four emotional support metrics are proposed to evaluate the mixed-initiative interactions. 
The analysis reveals the necessity and challenges of building mixed-initiative ESC systems. 
In the light of this, we propose a knowledge-enhanced mixed-initiative framework (KEMI) for ESC, which retrieves actual case knowledge from a large-scale mental health knowledge graph for generating mixed-initiative responses. 
Experimental results on two ESC datasets show the superiority of KEMI in both content-preserving evaluation and mixed initiative related analyses. 
\blfootnote{$^*$ The work described in this paper is substantially supported by a grant from the Research Grant Council of the Hong Kong Special Administrative Region, China (Project Code: 14200620).}
\blfootnote{$^\dagger$ Corresponding author.}
\end{abstract}

\section{Introduction}
As the world is making efforts to recover from Covid-19 and plans for future construction, emotional support is of great importance in resolving the widespread emotional distress and increased risk for psychiatric illness associated with the pandemic~\cite{covid19-mental-health,wsdm21-covid-analysis}. 
A wide range of emotional support conversation (ESC) systems are emerging to provide prompt and convenient emotional support for help-seekers, including mental health support~\cite{www21-mental-health,ICWSM22-mental-health}, counseling~\cite{tacl16-counseling,sigdial20-counseling,acl22-counseling} or motivational interviewing~\cite{mi-dataset,ijcnn21-motivate,tocss22-motivate}. 
Generally, the ESC system aims at reducing the user's emotional distress as well as assisting the user to identify and overcome the problem via conversations~\cite{esconv}. 

\begin{figure}
\centering
\includegraphics[width=0.48\textwidth]{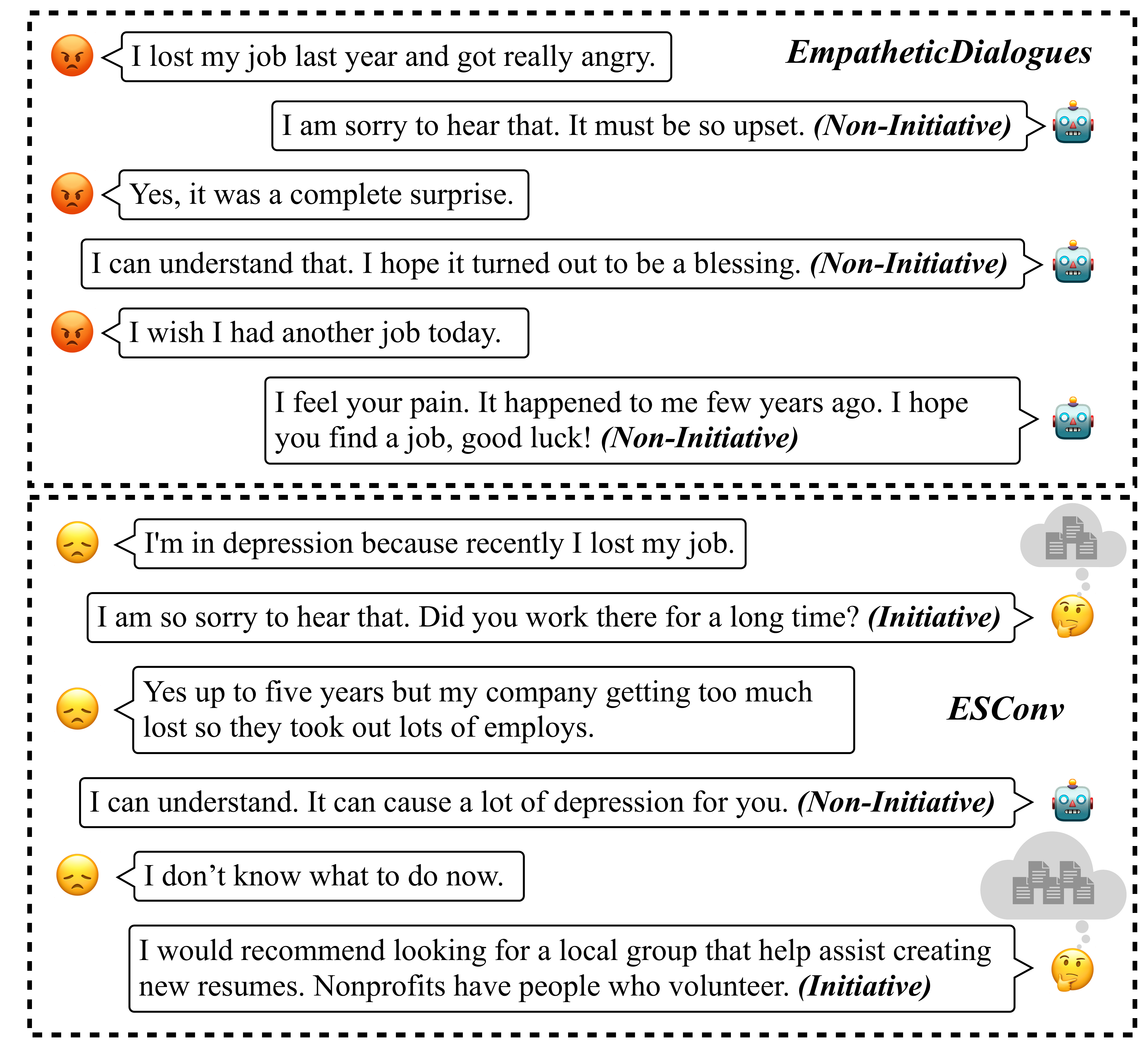}
\caption{Examples from \textsc{EmpatheticDialogues} and ESConv datasets with a similar job loss problem.}
\label{example}
\end{figure}

Mixed initiative is commonly defined as an intrinsic feature of human-AI interactions where the user and the system can both take the initiative in leading the interaction directions~\cite{allen1999mixed,umap20-proactive}. 
For example, mixed-initiative conversational information-seeking (CIS) systems~\cite{sigir19-clari,proactive_survey} can proactively initiate clarification interactions for resolving the ambiguity in the user query, instead of only reacting to the query. 
Accordingly, a mixed-initiative ESC system can proactively switch the initiative to provide an empathetic response or initiate a problem-solving discussion when appropriate. 
Many efforts have been made on the emotion reasoning for generating empathetic responses~\cite{sigdial20-counseling,acl20-counseling,emnlp22-esc,ijcai22-glhg}. 
Another line of work focuses on identifying the dialogue acts of the utterances~\cite{coling20-response-intent,wsdm22-da-counselling,acl22-question-intent} or predicting the next conversational strategies~\cite{eacl17-predict-da-mi,esconv,acl22-misc} in ESC systems.  
However, the feature of mixed initiative has not been investigated in existing ESC studies.

To facilitate the analysis on mixed-initiative ESC systems, we first propose an EAFR schema to annotate the utterances into different types with speaker roles and initiative types, named \textit{\textbf{E}xpression} (User-initiative), \textit{\textbf{A}ction} (Support-initiative), \textit{\textbf{F}eedback} (User Non-initiative), and \textit{\textbf{R}eflection} (System Non-initiative). 
Besides, four emotional support metrics are designed to measure the characteristics of initiative and non-initiative interactions in ESC, including \textit{Proactivity}, \textit{Information}, \textit{Repetition}, and \textit{Relaxation}. 

To analyze the necessity of considering mixed initiative in ESC systems, we conduct a preliminary analysis on the different interaction patterns between ESC and empathetic dialogues (ED). 
Firstly, the dialogue flow analysis shows that the system in ED generally serves as a passive role, while the system in ESC proactively switches the initiative role during the conversation. 
As shown in Figure~\ref{example}, the system in ED solely targets at comforting the user by reflecting their feelings or echoing their situations, \textit{i.e.}, \textit{Non-Initiative}. 
Differently, ESC systems are further expected to proactively explore the user's problem by asking clarifying questions and help the user overcome the problem by providing useful information or supportive suggestions, \textit{i.e.}, \textit{Initiative}. 
Furthermore, the analysis of the conversation progress and the emotional support metrics  reveal three challenges in building a mixed-initiative ESC system: \textit{1) When} should the system take the initiative during the conversation? \textit{2) What} kind of information is required for the system to initiate a subdialogue? \textit{3) How} could the system facilitate the mixed-initiative interactions?

According to these challenges, we define the problem of mixed-initiative ESC, which includes three sub-tasks: \textit{1) Strategy Prediction} to determine the mixed-initiative strategy in the next turn, \textit{2) Knowledge Selection} to collect the necessary knowledge for the next turn, and \textit{3) Response Generation} to produce emotional support responses with appropriate mixed-initiative strategy and knowledge. 
To tackle this problem, we propose a novel framework, named Knowledge Enhanced Mixed-Initiative model (KEMI), to build a mixed-initiative dialogue system for emotional support conversations with external domain-specific knowledge.  
In detail, KEMI first employs a knowledge acquisition module to acquire emotional support knowledge from a large-scale knowledge graph on mental health dialogues. Specifically, we expand the user utterance with generated commonsense knowledge as a query graph and then perform subgraph retrieval over the knowledge graph. 
Secondly, a response generation module conducts multi-task learning of strategy prediction and response generation in a sequence-to-sequence manner to generate mixed-initiative responses with external knowledge. 

The main contributions of this work are summarized as follows:
(1) To measure the mixed-initiative interactions in ESC systems, we propose an innovative analysis method, including an EAFR annotation schema and corresponding emotional support metrics. 
(2) We propose a novel knowledge-enhanced mixed-initiative framework for ESC, which retrieves external knowledge from mental health knowledge graph by subgraph retrieval using the query graph expanded with commonsense knowledge. 
(3) Experimental results show that the mixed initiative is of great importance in ESC, and the proposed method effectively outperforms existing methods on both content-preserving evaluation and mixed initiative analyses.

\section{Related Works}
\textbf{Emotional Support Conversation} ~
Similar to fine-grained sentiment analysis~\cite{absa-survey,acl21-absa,emnlp21-absa} in conversations~\cite{absa-dialog,absa-qa}, early works on emotional chatting mainly investigate approaches to detecting user emotions~\cite{dailydialog,aaai18-emo-chat} or incorporating emotional signals into response generation~\cite{cikm19-emo-chat,acl19-emo-chat}. 
As for empathetic dialogue systems~\cite{acl19-emp-dataset,emnlp21-emp-data}, evolving from emotion-aware response generation~\cite{emnlp19-moel,emnlp20-mime} and emotional style transfer~\cite{www21-mental-health}, more efforts have been made on emotional reasoning techniques~\cite{sigir21-emp-cause,emnlp21-emp-cause,emnlp21-finding-emp-cause,emnlp22-esc}.  
Some latest studies explore the utilization of external knowledge for enhancing the model capability of emotion reasoning, including commonsense knowledge graph~\cite{aaai21-emo-know,aaai22-emp-know}, generative commonsense model~\cite{aaai22-cem}, and domain-specific knowledge~\cite{sigdial20-counseling,acl22-counseling}. 
\citet{acl22-counseling} collectively exploit three kinds of external knowledge. 
Likewise, many ESC systems also leverage commonsense knowledge for response generation~\cite{acl22-misc,ijcai22-glhg}. 
However, the commonsense knowledge is rather abstractive without detailed information, so that it is less helpful for the ESC system to generate meaningful and informative responses. 
In this work, we employ the generative commonsense model for query expansion to retrieve actual case knowledge from an external knowledge graph.

\begin{table*}[!t]
    \centering
\fontsize{9}{10.5}\selectfont
    \begin{adjustbox}{max width=\textwidth}
    \begin{tabular}{lllp{6cm}p{5.5cm}}
    \toprule
    Role & Type & EAFR & Definition & Sample Utterances \\
    \midrule
    \multirow{2}{*}{User} & \multirow{2}{*}{Initiative} & \multirow{2}{*}{Expression} & The user describes details or expresses feelings & My school was closed due to the pandemic. \\
    &&&  about the situation.& I feel so frustrated. \\
    \midrule
    &&&The system requests for information related to& How are your feelings at that time?\\
    System & Initiative & Action &  the problem or provides suggestions and infor- & Deep breaths can help people calm down. \\
    &&&mation for helping the user solve the problem. & Some researches has found that ...\\
    \midrule
    \multirow{2}{*}{User} & \multirow{2}{*}{Non-Initiative} & \multirow{2}{*}{Feedback} & The user responds to the system’s request or  & Okay, this makes me feel better.\\
    &&&delivers opinions on the system's statement. & No, I haven't.\\
    \midrule
    &&&The system conveys the empathy to the user’s& I understand you. I would also have been \\
    System & Non-Initiative & Reflection &  emotion or shares similar experiences and  & really frustrated if that happened to me.\\
    &&&feelings to comfort the user. & I'm sorry to hear about that. \\
    \bottomrule
    \end{tabular}
    \end{adjustbox}
    \caption{Definition and Examples for EAFR Schema Reflecting Patterns of Initiative Switch between Dialogue Participants in Emotional Support Conversations.}
    \label{tab:eafr}
\end{table*}

\noindent\textbf{Mixed-initiative Dialogue} ~
Recent years have witnessed many efforts on developing mixed-initiative conversational systems for various dialogues, such as information-seeking dialogues~\cite{www20-clari,sigir19-clari}, open-domain dialogues~\cite{acl19-proactive,sigdial21-mix-topic,sigir22-proactive}, recommendation dialogues~\cite{unicorn}, conversational question answering~\cite{pacific}. 
Despite the importance of mixed initiative in ESC systems, this area has not been investigated. 
One closely related research scope is to recognize the conversation strategies~\cite{esconv,eacl17-predict-da-mi} or the dialogue acts~\cite{wsdm22-da-counselling,coling20-response-intent,acl22-question-intent,www22-use} of the utterances in ESC systems. 
However, these studies only focus on predicting the support strategies, instead of actually involving mixed-initiative interactions in ESC.

In addition, measuring mixed initiative is also regarded as an essential perspective for assessing dialogue quality~\cite{tois21-mix,sigir20-mix-analysis,qrfa}. 
Due to the high expenses in human evaluation, \citet{wsdm22-mix-usersim} and \citet{kdd20-eval-crs} investigate user simulation for evaluating the mixed-initiative interactions in conversational systems. 
In this work, we investigate several metrics for measuring the characteristics of the mixed initiative in ESC systems.

\section{Preliminary Analysis}
\subsection{EAFR Schema \&  Metrics}\label{sec:eafr}
Inspired by the ConversationShape~\cite{tois21-mix} for the analysis of mixed-initiative CIS systems, we first propose an EAFR annotation schema to study the mixed initiative in ESC systems. The EAFR annotation schema classifies the utterance in ESC into four categories w.r.t the role of speakers and the type of initiative, including \textit{\textbf{Expression}} (User-initiative), \textit{\textbf{Action}} (System-initiative), \textit{\textbf{Feedback}} (User Non-Initiative), and \textit{\textbf{Reflection}} (System Non-Initiative). 
Definitions and examples of each type are presented in Table~\ref{tab:eafr}.

Then, each utterance $i$ in a dialogue is annotated as a tuple $(r_{i}, t_{i}, \bm{v}_{i}, e_{i})$ for analysis. $r_{i}\in \{\text{User}(U), \text{System}(S)\}$ denotes the speaker role. $t_{i}\in \{\text{Initiative}(I), \text{Non-Initiative}(N)\}$ denotes the initiative type.  $\bm{v}_{i}\in \{0,1\}^{|V|}$ denotes the one-hot vocabulary embeddings. $e_{i}\in [1,5]$ denotes the level of emotion intensity\footnote{A decrease
from the intensity reflects
emotion improvement~\cite{esconv}.}. 
We further design four emotional support metrics for investigating patterns of mixed initiative in ESC systems as follows:
\begin{itemize}[leftmargin=*]
    \item \textbf{Proactivity}: how proactive is the system in the emotional support conversation? 
    \begin{equation}\small
        \text{Pro} = \frac{1}{\sum_{i=1}^{n}\mathcal{I}(r_{i}=S)}\sum\nolimits_{i=1}^{n}\mathcal{I}(r_{i}=S,t_{i}=I)
    \end{equation}
    denotes the ratio of system-initiative interactions.  
    \item \textbf{Information}: how much information does the system contribute to the dialogue?
    \begin{equation}\small
        \text{Inf} = \frac{\sum_{i=1}^{n}\sum_{k=1}^{|V|}\mathcal{I}(r_{i} = S, v_{ik}=1, \sum_{j=1}^{i-1} v_{jk}=0)}{\sum_{i=1}^{n}\mathcal{I}(r_{i}=S)}
    \end{equation}
    represents the average number of new frequent terms\footnote{We only consider frequent terms that appear in the dialogue more than once. Standard pre-processing pipeline is adopted: remove punctuation, tokenization, lowercase, remove stopwords, and apply the English Snowball stemmer.} that are introduced by the system.
    \item \textbf{Repetition}: how often does the system follow up on the topic introduced by the user?  \vspace{-0.2cm}
    \begin{equation}\small
        \text{Rep} = \frac{\sum\limits_{i=1}^{n}\sum\limits_{k=1}^{|V|}\mathcal{I}(r_{i}=S, v_{ik}=1, \sum\limits_{j=1}^{i-1} v_{jk}[r_{j}=U]>0)}{\sum_{i=1}^{n}\mathcal{I}(r_{i}=S)}
    \end{equation}
    represents the average number of repeated frequent terms that are introduced by the user and mentioned by the system.
    \item \textbf{Relaxation}: how well does the system relax the emotional intensity of the user?
    \begin{equation}\small\label{eq:rel}
        \text{Rel}_{i}[r_{i}=S] = e_{<i}[r_{<i}=U] - e_{>i}[r_{>i}=U]
    \end{equation}
    \begin{equation}\small
        \text{Rel} = \frac{1}{\sum_{i=1}^{n}\mathcal{I}(r_{i}=S)}\sum\nolimits_{i=1}^{n} \text{Rel}_{i}[r_{i}=S]
    \end{equation}
    represents the change of the user's emotion intensity. 
    $e_{<i}[r_{<i}=U]$ and $e_{>i}[r_{>i}=U]$ denote the emotion intensity of the first user utterance before and after the utterance $i$, respectively.  
\end{itemize}

\subsection{Analysis of Mixed Initiative in ESC}\label{sec:preliminary}
To reveal the necessity of incorporating mixed initiative into ESC systems, we analyze the different interaction patterns between empathetic dialogues (ED) and emotional support conversations (ESC): 
(i) \textsc{EmpatheticDialogues}~\cite{acl19-emp-dataset}, a dataset for ED that aims to provide empathetic responses for comforting the help-seeker, and (ii) ESConv~\cite{esconv}, a dataset for ESC that aims to not only reduce users’ emotional distress, but also help them understand and overcome the issues they face. 

Due to the space limitation, we present the detailed analysis in Appendix~\ref{app:preliminary}, including (i) the visualization of dialogue flow
that indicates the initiative patterns between the user and system (\ref{app:dial_flow}); (ii) the visualization of conversation progress that shows the phased change of the user's emotion intensity (\ref{app:conv_prog}); and (iii) the evaluation of emotional support metrics that quantifies different aspects of mixed-initiative interactions (\ref{app:metrics}).

\subsection{Challenges of Mixed Initiative in ESC}\label{sec:taxonomy}
The preliminary analysis reveals the importance of mixed-initiative interactions in ESC systems. Meanwhile, it is also challenging to balance the mixed-initiative interactions, as overacting in one way or taking the initiative inappropriately can be harmful to the emotional support conversations. 
Based on these analyses, we identify three key challenges in building a mixed-initiative ESC system:

\noindent\textbf{\textit{1) When} should the system take the initiative during the conversation?}
The analysis of conversation progress (\ref{app:conv_prog}) shows that taking initiative at different phases of the conversation may lead to different impacts on the user's emotional state. In particular, support strategies or dialogue acts attach great importance to conversational effectiveness in ESC~\cite{acl20-counseling,acl22-misc}. 
Therefore, it is a crucial capability for the ESC system to determine whether to take the initiative at each conversation turn.

\noindent\textbf{\textit{2) What} kind of information is required for the system to initiate a subdialogue?}
The analysis of mixed initiative metrics (\ref{app:metrics}) show that the initiative system utterances are much informative than the non-initiative ones. 
Therefore, it is of great importance to discover necessary information and knowledge to make an appropriate mixed-initiative interaction. 
Researchers~\cite{esskills} in communication and sociology states that the helpfulness of supportive statement is contingent on the following knowledge: 
(i) \textit{Affective} Knowledge, the emotion recognition of the user's affective state, 
(ii) \textit{Causal} Knowledge, the emotional reasoning of stressors that cause the current affective state of the user, and
(iii) \textit{Cognitive} Knowledge, the cognitive analysis of coping processes to solve the core problematic situation that the user faces. 
    
\noindent\textbf{\textit{3) How} could the system facilitate the mixed-initiative interactions? }
Since the system in ESC ultimately provides a natural language utterance to interact with the user, this challenge can be defined as a function that generates an initiative-aware utterance based on the given information. 

\subsection{Problem Definition}
Similar to the ED problem, the ESC problem is typically defined as: given the dialogue context $\mathcal{C} = \{u_1, u_2, ..., u_{t}\}$ and the description of the user's problematic situation $s$, the goal is to estimate a function $p(r|\mathcal{C},s)$ that generates the target response $r$. 
In the light of the challenges discussed in Section~\ref{sec:taxonomy}, we further define the mixed-initiative emotion support conversation problem with the following three sub-tasks, corresponding to the above three challenges:  

\noindent 1) \textit{Strategy Prediction} predicts the support strategy $y$ that can be regarded as the fine-grained initiative. 

\noindent 2) \textit{Knowledge Selection} selects appropriate knowledge $k$ from the available resources $\mathcal{K}$. 

\noindent 3) \textit{Response Generation} generates the mixed-initiative response $r$ based on the predicted strategy and the selected knowledge.

\section{Method}
\begin{figure*}
\centering
\includegraphics[width=0.95\textwidth]{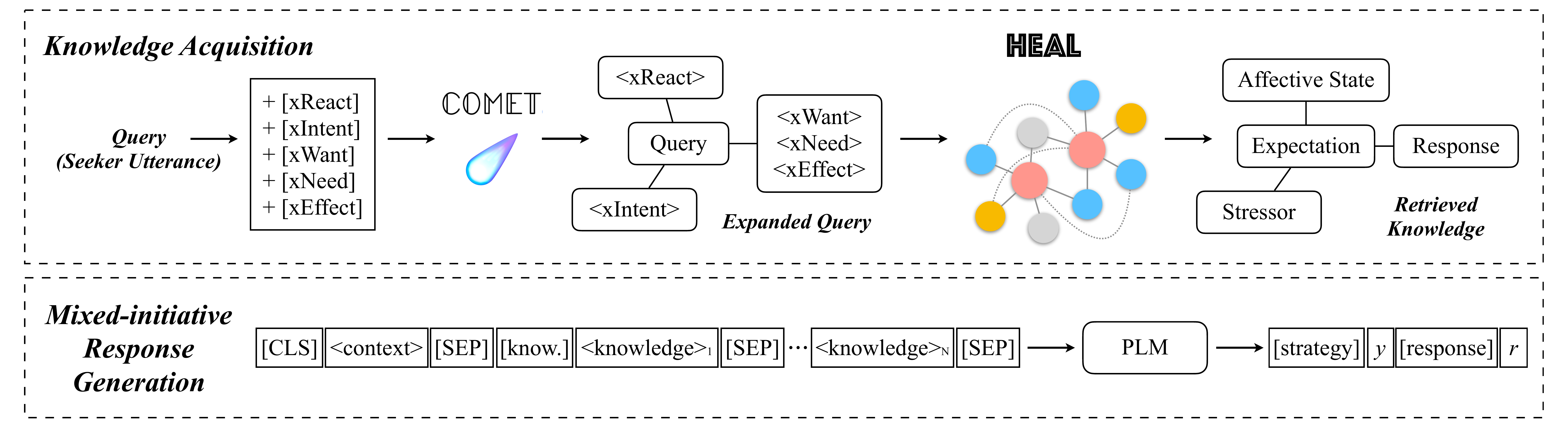}
\caption{Overview of KEMI. Each expanded query is represented as a graph to retrieve subgraphs from  \texttt{HEAL}, and each subgraph in \texttt{HEAL} can be regarded as an actual case of emotional support conversations.}
\label{kemi}
\end{figure*}

Motivated by the analysis in the last section, we propose the KEMI framework that aims to generate mixed-initiative responses with external knowledge. 
As illustrated in Figure~\ref{kemi}, KEMI contains two parts: 1) Knowledge Acquisition, and 2) Mixed-initiative Response Generation.

\subsection{Knowledge Acquisition}
Commonsense knowledge is widely adopted to enhance the emotion reasoning in ESC systems. Despite the wide usage of commonsense knowledge in ESC systems, it is usually succinct and lacks specific context information. 
We propose an approach to retrieve relevant actual cases of ESC from a large-scale mental health knowledge graph, namely \texttt{HEAL}~\cite{heal}, for compensating the deficiency of commonsense knowledge.

\subsubsection{Query Expansion with COMET}
Given the user utterance $u_t$ at the current turn $t$, a straight-forward knowledge acquisition approach is to use $u_t$ as the query to directly retrieve actual cases from the \texttt{HEAL} KG. 
However, there is limited information provided by the user utterance, which may hinder the preciseness and explainability of the knowledge retrieval. 
To this end, we exploit \texttt{COMET}~\cite{comet}, a commonsense knowledge generator, to expand the query with multi-perspective additional information regarding the user's affective and cognitive state. 

Specifically, the current user utterance $u_t$ is fed into \texttt{COMET} with five special relation tokens,  $p\in\{\texttt{[xReact]}, \texttt{[xIntent]}, \texttt{[xWant]}, \texttt{[xNeed]}\\, \texttt{[xEffect]}\}$, to generate commonsense inference $c_p$ for the relation $p$, \textit{i.e.}, $c_p = \texttt{COMET}(p, u_t)$.  
Definitions of each commonsense relation can be found in Appendix~\ref{app:comet}. 
Then the original user utterance $u_t$ can be expanded with commonsense knowledge $\{c_p\}$.  

\subsubsection{Query Graph Construction}
The actual case in HEAL~\cite{heal} is represented as a graph structure. 
Specifically, we consider 4 out of 5 types of nodes in \texttt{HEAL} that are related to response generation: \textit{1) expectation}: commonly asked questions by the user in an emotional support conversation; 
\textit{2) affective state}: emotional states associated with each speaker; 
\textit{3) stressor}: the cause of emotional issues;  
and \textit{4) response}: frequent types of responses by the system to address the user's problems.
Edges are constructed to build the connections between nodes according to actual emotional support conversations. 
More details of \texttt{HEAL} can be found in Appendix~\ref{app:heal}. 

In accordance with the \texttt{HEAL} knowledge graph, the relation \texttt{[xReact]}, which reveals the user's emotional state, provides the same information as nodes in \texttt{HEAL} with the type of \textit{affective state}. 
The relation \texttt{[xIntent]}, which reveals the causes of the user's current situation, also shares the same information as nodes in \texttt{HEAL} with the type of \textit{stressor}. 
The rest of relations, including \texttt{[xWant]}, \texttt{[xNeed]}, and \texttt{[xEffect]}, which reveal the user's cognitive state, are relevant to the \textit{responses} for addressing the user's problem.  
Therefore, the expanded query $\hat{u}_t=\{u_t,\{c_p\}\}$ can be represented as a graph with abstractive entity descriptions, as shown in Figure~\ref{kemi}.

\subsubsection{Subgraph Retrieval}  
To avoid enumerating all the subgraphs in \texttt{HEAL}, which is a densely-connected graph (over 2 million subgraphs), we propose a subgraph retrieval approach to select the top relevant subgraphs to form a candidate set. 
We first retrieve top-$K$ entities relevant to each abstractive entity description in the expanded query graph $\hat{u}_t$. 
Specifically, we use sentence-BERT~\cite{sbert} to be an embedding-based retriever $f_r(\cdot)$ for modeling the semantic similarity between the entities in the query and \texttt{HEAL}. 
With the retrieved top-$K$ entities for each type of nodes, we merge them based on the edge connections in the knowledge graph to induce candidate subgraphs. 
Finally, we adopt top-$N$ candidate subgraphs as the retrieved knowledge $\mathcal{K}$. The subgraphs are ranked by the sum of similarity scores of each node in the subgraph $E=\{e_\text{exp},e_\text{aff},e_\text{str},e_\text{resp}\}$:
\begin{equation}\small\begin{aligned}
    \textbf{Sim}(\hat{u}_t, E) =& f_r(u_t,e_\text{exp})+f_r(c_\text{xR},e_\text{aff})+f_r(c_\text{xI},e_\text{str})\\
    &+f_r([c_\text{xW},c_\text{xN},c_\text{xE}],e_\text{resp}).
\end{aligned}
\end{equation}

\subsection{Mixed-initiative Response Generation}\label{sec:mtl}
Given the dialogue context $\mathcal{C}$ and the retrieved knowledge $\mathcal{K}$, we first encode them into distributed representations with contextualized encoders. Specifically, we add special tokens to differentiate the roles of user and system as well as different types of knowledge as: 
\begin{equation*}\small
    \texttt{<context>} = \texttt{[situ.]},s, \texttt{[usr]},u_1,\texttt{[sys]},u_2,...
\end{equation*}
\begin{equation*}\small
    \texttt{<know.>} = \texttt{[xR.]},c_\text{xR},\texttt{[xI.]},...,\texttt{[Aff.]},e_\text{aff},...
\end{equation*}

Pretrained language models (PLMs), \textit{e.g.}, GPT2~\cite{gpt2}, have shown superior capability of generating high-quality responses in many dialogue systems, especially those PLMs pretrained on dialogue corpus, \textit{e.g.}, BlenderBot~\cite{blenderbot}. 
To leverage the advantages of these generative PLMs, we reformulate the mixed-initiative emotional support conversation problem as a Seq2Seq problem, which linearizes the input and output as a sequence of tokens as follows: 
\begin{equation*}\small
    X=\texttt{[CLS]}, \texttt{<context>}, \texttt{[know.]}, \texttt{<know.>}_i, ...
\end{equation*}\begin{equation*}\small
    Y=\texttt{[strategy]}, y, \texttt{[response]}, r 
\end{equation*}
where $X$ and $Y$ are the linearized input and output sequences for Seq2Seq learning. Then the model is trained to maximize the negative log likelihood: 
\begin{equation}\small
    \mathcal{L} = -\frac{1}{L}\sum\nolimits^{L}_{l=1}\log P(Y_{l}|Y_{<l};X).
\end{equation}

\section{Experiment}
\subsection{Experimental Setups}
\textbf{Datasets} ~ 
We adopt the following two datasets for the evaluation: (i) ESConv~\cite{esconv}, an emotional support conversation dataset, contains 1,300 dialogues with 38,365 utterances and 8 types of support strategies. We adopt the original train/dev/test split; and (ii) MI~\cite{mi-dataset}, a motivational interviewing dataset, contains 284 counseling sessions with 22,719 utterances and 10 types of behavior strategies. We randomly split the dataset for train/dev/test by 8:1:1\footnote{Since there is no speaker label in the MI dataset, it is only adopted for response generation evaluation while the analysis of mixed initiative is not applicable.}. 

\noindent \textbf{Evaluation Metrics} ~ 
As for automatic evaluation, we adopt Macro F1 as the strategy prediction metric. Following previous studies~\cite{esconv,acl22-misc}, Perplexity (PPL), BLEU-$n$ (B-$n$), and ROUGE-L (R-L) are included for the evaluation of response generation.

\noindent \textbf{Baselines} ~ 
We provide extensive comparisons with both non-PLM and PLM-based methods, including three Transformer-based methods (Transformer~\cite{transformer}, MoEL~\cite{emnlp19-moel}, and MIME~\cite{emnlp20-mime}) and four BlenderBot-based methods (BlenderBot~\cite{blenderbot}, BlenderBot-Joint~\cite{esconv}, GLHG~\cite{ijcai22-glhg}\footnote{Since GLHG leverages the problem type as an additional label, we also report the ablation result for a fair comparison, \textit{i.e.}, GLHG w/o $\mathcal{L}_2$ Loss.} , and MISC~\cite{acl22-misc}\footnote{Due to a different train/test split adopted in \citet{acl22-misc}, we reproduce the performance of MISC on the standard split of ESConv~\cite{esconv}.}). 
Details about these baselines can be found in Appendix~\ref{app:baseline}.

\noindent \textbf{Implementation Details} ~KEMI is based on the BlenderBot model~\cite{blenderbot}. 
Following previous BlenderBot-based models~\cite{esconv,ijcai22-glhg,acl22-misc}, we adopt the small version\footnote{\url{https://huggingface.co/facebook/blenderbot_small-90M}} of BlenderBot in experiments. 
The learning rate and the warmup step are set to be 3e-5 and 100, respectively. The max input sequence length and the max target sequence length are 160 and 40, respectively. 
We retrieve the top-$1$ subgraph from \texttt{HEAL} as the knowledge. 
The training epoch is set to 5 and the best model is saved according to the PPL score in the dev set.\footnote{\url{https://github.com/dengyang17/KEMI}}

\subsection{Overall Performance}
Table~\ref{tab:esconv_res} and Table~\ref{tab:mi_res} summarize the experimental results on the ESConv and MI dataset, respectively. 
Among the baselines, BlenderBot-based methods largely outperform Transformer-based methods by leveraging the valuable pretrained knowledge. 
GLHG and MISC effectively exploit the commonsense knowledge to improve the performance of response generation. 
Besides, the joint learning with strategy prediction task is beneficial to the performance of response generation. 
Finally, KEMI substantially outperforms other methods with a noticeable margin. 
This indicates the domain-specific actual case knowledge from \texttt{HEAL} can alleviate the reliance on large-scale PLMs. 
Compared with commonsense knowledge, the  knowledge from \texttt{HEAL} is much more effective in predicting support strategies, as this relevant knowledge can serve as an real example for guiding the system to respond.

\begin{table}
    \centering
    \begin{adjustbox}{max width=0.48\textwidth}
    \setlength{\tabcolsep}{1.2mm}{
    \begin{tabular}{lccccc}
    \toprule
    Model & F1$\uparrow$ & PPL$\downarrow$ &  B-2$\uparrow$ & B-4$\uparrow$  &R-L$\uparrow$ \\
    \midrule
       Transformer$^*$~\cite{transformer} & -&  81.55&5.66&1.31&14.68 \\
       MoEL$^*$~\cite{emnlp19-moel} &-& 62.93&5.02&1.14&14.21\\
       MIME$^*$~\cite{emnlp20-mime} &-& 43.27&4.82&1.03&14.83 \\
       BlenderBot$^{**}$~\cite{blenderbot} &-& 16.23&5.45&-&15.43\\
       GLHG$^*$~\cite{ijcai22-glhg} &-& \underline{\textbf{15.67}}&7.57&2.13&16.37 \\
       GLHG w/o $\mathcal{L}_2$ Loss$^*$~\cite{ijcai22-glhg}&-&-&6.15&1.75&15.87\\
       BlenderBot-Joint~\cite{esconv} &19.23& 16.15&5.52&1.29&15.51\\
       MISC~\cite{acl22-misc} &\underline{19.89}& 16.08 & \underline{7.62}&\underline{2.19}&\underline{16.40}\\
       \midrule
       KEMI  & \textbf{24.66}$^\dagger$ & 15.92& \textbf{8.31}$^\dagger$ & \textbf{2.51}$^\dagger$ & \textbf{17.05}$^\dagger$ \\
    \bottomrule
    \end{tabular}}
    \end{adjustbox}
    \caption{Experimental results on ESConv. $^*$ and $^{**}$ indicate the results reported in \citet{ijcai22-glhg} and \citet{esconv} respectively. Other results are reproduced. $^\dagger$ indicates statistically significant improvement ($p$<0.05) over \underline{the best baseline}.}
    \label{tab:esconv_res}
\end{table}

\begin{table}
    \centering
    \begin{adjustbox}{max width=0.48\textwidth}
    \setlength{\tabcolsep}{1.2mm}{
    \begin{tabular}{lccccccc}
    \toprule
    Model & F1$\uparrow$ & PPL$\downarrow$ &B-2$\uparrow$ & B-4$\uparrow$  &R-L$\uparrow$ \\
    \midrule
    Transformer~\cite{transformer}& -&65.52&6.23&1.52&15.04\\
       BlenderBot~\cite{blenderbot}&-&16.06&6.57&1.66&15.64\\
       BlenderBot-Joint~\cite{esconv} &22.66&14.74&7.28&2.18&16.41\\
       MISC~\cite{acl22-misc} &\underline{22.68}&\underline{14.33}&\underline{7.75}&\underline{2.30}&\underline{17.11} \\
       \midrule
       KEMI  & \textbf{25.91}$^\dagger$&\textbf{13.84}$^\dagger$&\textbf{8.52}$^\dagger$&\textbf{2.72}$^\dagger$&\textbf{18.00}$^\dagger$\\
    \bottomrule
    \end{tabular}}
    \end{adjustbox}
    \caption{Experimental results on MI Counseling. }
    \label{tab:mi_res}
\end{table}

\subsection{Human Evaluation}
Following previous studies~\cite{esconv,ijcai22-glhg}, we conduct human evaluation to compare the generated responses from two given models on five aspects: 
1) \textit{Fluency}: which model's response is more fluent? 
2) \textit{Identification}: which model's response is more skillful in identifying the user's problem? 
3) \textit{Comforting}: which model's response is better at comforting the user?
4) \textit{Suggestion}: which model can give more helpful and informative suggestions?  
5) \textit{Overall}: which model's response is generally better? 
We randomly sample 100 dialogues from ESConv and three annotators are asked to determine the \textit{Win/Tie/Lose} for each comparison.  

\begin{table}
    \centering
    \setlength{\tabcolsep}{1.5mm}{
    \begin{tabular}{lccccccccc}
    \toprule
       \multirow{2}{*}{ vs.}& \multicolumn{3}{c}{BlenderBot-Joint}& \multicolumn{3}{c}{MISC}\\
       \cmidrule(lr){2-4}\cmidrule(lr){5-7}
        & Win & Tie & Loss & Win & Tie & Loss  \\
       \midrule
        Flu.  & 26\% & \textbf{51\%} & 23\% & 37\% & \textbf{47\%} & 16\%\\
        Ide.  & \textbf{50\%} & 38\% & 12\% & \textbf{46\%} &30\% &24\% \\
        Com.  & \textbf{46\%} &40\% &14\% &\textbf{44\%} &30\% &26\% \\
        Sug.  & \textbf{52\%} &22\% &26\% &\textbf{52\%} &16\% &28\% \\
        Ove.  & \textbf{62\%} &20\% &18\% & \textbf{70\%} &12\% &18\% \\
        \bottomrule
    \end{tabular}}
    \caption{Human evaluation results (KEMI vs.).}
    \label{tab:human_eval}
\end{table}

Table~\ref{tab:human_eval} presents the human evaluation results. We compare the generated responses from KEMI with those produced by other two baselines, BlenderBot-Joint and MISC. 
The results show that KEMI achieves remarkable improvement on initiative interactions, including \textit{Identification} and \textit{Suggestion}. 
Consequently, KEMI can generate more satisfactory and helpful responses than other methods, according to the \textit{Overall} metric.

\subsection{Ablation Study}
In order to investigate the effect of each sub-task and each type of knowledge on the final performance, we report the experimental results of the ablation study in Table~\ref{tab:ablation}. 
In general, both the strategy prediction and the knowledge selection tasks as well as all types of knowledge contribute to the final performance more or less. 
There are several notable observations in detailed comparisons:   
(i) The knowledge from \texttt{HEAL} is the key to the improvement on the strategy prediction task, since the actual case knowledge can provide a good guidance for the next support strategy.  
(ii) Different from discarding the actual case knowledge (w/o \texttt{HEAL}), discarding the commonsense knowledge (w/o \texttt{COMET}) brings a positive effect on the fluency metrics (PPL), as the commonsense knowledge is not a natural sentence. However, the \texttt{COMET} contributes more on the content-preserving metrics (BLEU and ROUGE) than the \texttt{HEAL}, indicating that the succinct commonsense knowledge can be more precise. 
(iii) Among the three types of knowledge, cognitive knowledge is the most effective one for both strategy prediction and response generation tasks. 
(iv) Using Oracle strategy and Oracle knowledge substantially improves the overall performance, which demonstrates the effectiveness of considering these two sub-tasks in ESC systems. The performance gap between KEMI and Oracle also shows that the knowledge selection is very challenging and there is still much room for improvement. 

\begin{table}
    \centering
    \begin{adjustbox}{max width=0.48\textwidth}
    \begin{tabular}{llcccc}
    \toprule
      Strategy & Knowledge & F1$\uparrow$ & PPL$\downarrow$ &B-2$\uparrow$   &R-L$\uparrow$ \\
      \midrule
      - & -&- & 16.23&5.45&15.43 \\
       - & KEMI & - &16.16&6.54&16.21\\
       \midrule
      Joint & KEMI&24.66 & 15.92& 8.31  & 17.05\\
      Joint & w/o \texttt{COMET}&23.26&15.74&7.60&16.47\\
      Joint & w/o \texttt{HEAL}&19.99& 16.08 & 7.98&16.92\\
      Joint & w/o \textit{Affective} & 22.68&16.08&8.22&16.98\\
      Joint & w/o \textit{Causal} &23.14&15.94&8.16&16.92\\
      Joint & w/o \textit{Cognitive} & 20.24&16.22&7.62&16.64\\
      Joint & Oracle& \textbf{32.38}&12.79&18.45&28.01\\
      \midrule
      Oracle & KEMI& -& 15.92&9.75&18.81 \\
      Oracle & Oracle& - &\textbf{12.78}&\textbf{19.11}&\textbf{28.88}\\
      \bottomrule
    \end{tabular}
    \end{adjustbox}
    \caption{Ablation study. Oracle knowledge is obtained by the lexical match between the reference response and the candidate knowledge from \texttt{HEAL}.}
    \label{tab:ablation}
\end{table}

\begin{table}
    \centering
    \begin{adjustbox}{max width=0.48\textwidth}
    \setlength{\tabcolsep}{0.5mm}{
    \begin{tabular}{l|cc|ccc|ccc|ccc}
    \toprule
    \multicolumn{1}{c}{}     
    & \multicolumn{2}{c}{Proactivity} &\multicolumn{3}{c}{Information}&\multicolumn{3}{c}{Repetition} &\multicolumn{3}{c}{Relaxation} \\
    \cmidrule(lr){2-3}\cmidrule(lr){4-6}\cmidrule(lr){7-9}\cmidrule(lr){10-12}
    
    \multicolumn{1}{c}{} & Init. & \multicolumn{1}{c}{Non.}  & Init. & Non. & \multicolumn{1}{c}{All}& Init. & Non. & \multicolumn{1}{c}{All}& Init. & Non. & \multicolumn{1}{c}{All} \\
    \midrule
    BB & 0.36&\textbf{0.64}&1.79&1.32&1.48&1.00&1.11&1.07&-0.01&0.11&0.07\\ 
    BB-J & \textbf{0.68} & 0.32  &1.89&1.18&1.66 &1.18 &1.09&\textbf{1.15} &0.01&0.07&0.03 \\
    MISC 
    & 0.61&0.39 &1.91&1.25&1.65 &1.16&\textbf{1.12}&1.14&0.00&0.04&0.02\\
    KEMI &0.45 & 0.55  &\textbf{2.04}&\textbf{1.40}&\textbf{1.68} &\textbf{1.18}&1.09&1.13&\textbf{0.09}&\textbf{0.13}&\textbf{0.11} \\
    \midrule
    REF & 0.51 & 0.49 & 3.09 & 3.01 & 3.05& 1.12 & 1.06 & 1.09& 0.10 & 0.13 & 0.11\\
    \bottomrule
    \end{tabular}}
    \end{adjustbox}
    \caption{Emotional support metrics. BB and BB-J denote BlenderBot and BlenderBot-Joint.}
    \label{exp:metrics}
\end{table}

\subsection{Analysis of Mixed Initiative}
We conduct the mixed initiative analysis introduced in Section~\ref{sec:preliminary} over the proposed KEMI method and other baselines. 
Since the calculation of the Relaxation metric in Eq.(\ref{eq:rel}) requires the emotion intensity score of the user feedback, we adopt a model-based user simulator for automatic evaluation, which is described in Appendix~\ref{app:usersim}. 

\subsubsection{Emotional Support Metrics}

\begin{figure*}
\centering
\includegraphics[width=\textwidth]{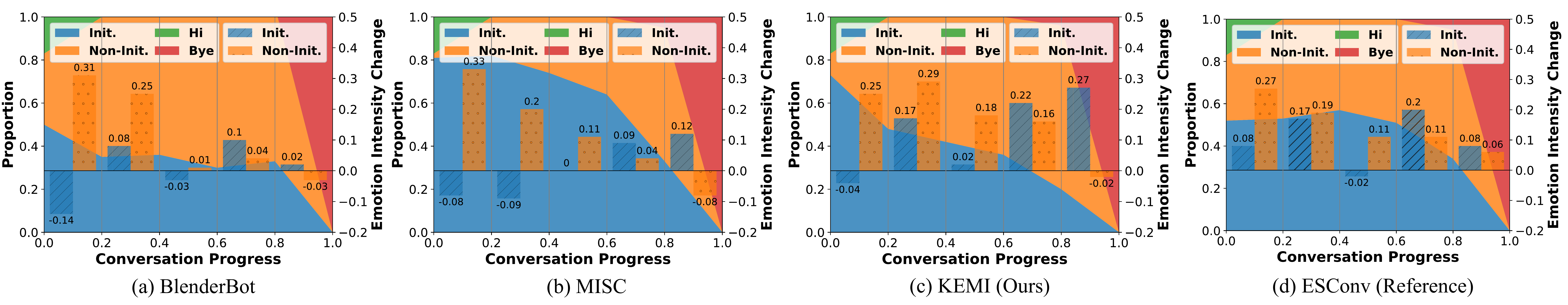}
\caption{The distribution of system utterance initiative (the stack plot) and the user's emotion intensity change (the bar chart) at different conversation progress. Higher scores of the emotion intensity change represent better emotion improvement of users.}
\label{fig:conv_prog}
\end{figure*}

\begin{figure*}
\centering
\includegraphics[width=\textwidth]{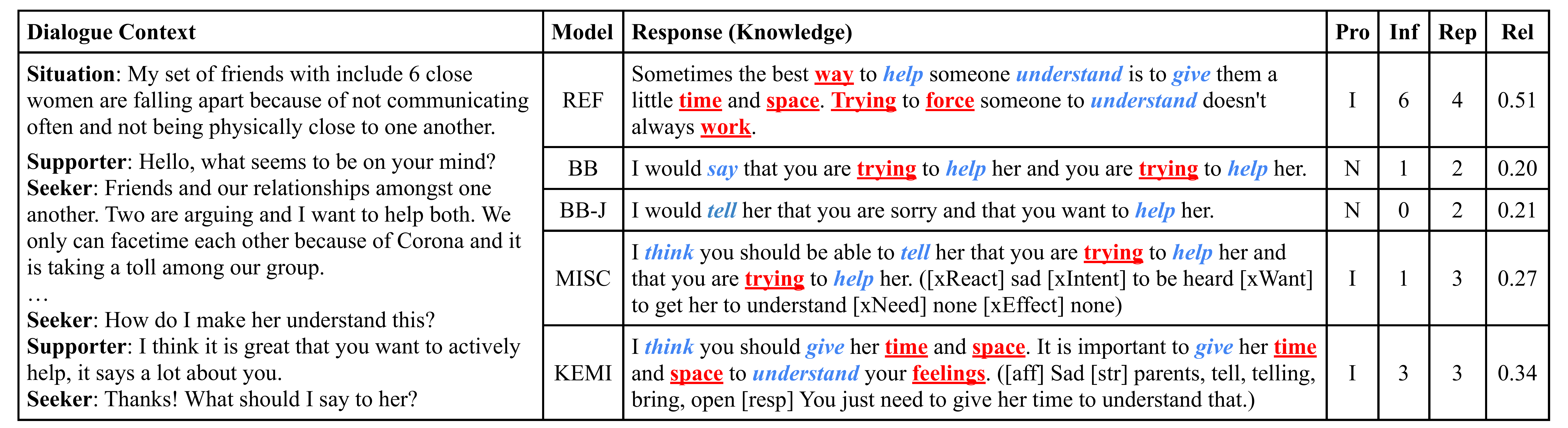}
\caption{Case study. \textbf{Bold} terms denote \underline{new} (red) and \textit{repeated} (blue) frequent terms respectively.}
\label{fig:case}
\end{figure*}

Table~\ref{exp:metrics} summarizes the results of the four emotional support metrics for the generated responses from four BlenderBot-based methods and the reference responses in the test set. Note that, for a fair comparison, we also adopt Eq.(\ref{eq:sim_rel}) to calculate the Relaxation metric for the reference responses in the test set (\textit{i.e.}, REF). 
It can be observed that (i) As for the Proactivity metric, BlenderBot tends to act passively in ESC. While BlenderBot-Joint and MISC overly take the initiative after simply taking into account the support strategies. KEMI effectively balances the proportion of initiative and non-initiative interactions in ESC. (ii) With the actual case knowledge, KEMI can generate much informative responses than other baselines w.r.t the Information metric. However, there is still a large gap to reach the reference responses. (iii) Indeed, it is relatively easier to generate responses that repeat the previous information w.r.t the Repetition metric. (iv) KEMI outperforms other baselines in terms of the Relaxation metric on the initiative interactions with a large margin, which shows the superiority of KEMI on taking the initiative role for helping the user to solve emotional problems.

\subsubsection{Conversation Progress}
We conduct the conversation progress analysis by dividing the whole conversation into five equal length intervals and observing the change of users' emotion intensity levels at each conversation phase. 
As the results shown in Figure~\ref{fig:conv_prog}, we observe that BlenderBot and MISC have a clear inclination to take non-initiative and initiative interactions in all stages of the conversation, respectively. Our KEMI method shares a more similar progress as the reference conversation with a balanced interaction pattern. 
More importantly, the initiative responses generated by KEMI has a more positive impact on the user's emotional intensity than other baselines, especially in the last two stages of the conversation. 
This result indicates that KEMI effectively takes the initiative to generate responses that can provide suggestions or information for relaxing the help-seekers by solving their emotional problems.

\vspace{-0.05cm}

\subsection{Case Study}
To intuitively show the superiority of KEMI over other baselines, Figure~\ref{fig:case} presents a case study of generated responses with the scores of mixed-initiative metrics. 
In the reference response, the system takes the initiative to provide useful suggestions to the user for solving her/his problem, which effectively reduce the user's emotional intensity. 
Among the generated responses, BlenderBot and BlenderBot-Joint decide to convey empathy to the user by paraphrasing the previous information, while MISC and KEMI proactively initiate a discussion about potential solutions to the problem. Based on the Relaxation metric, two initiative responses can better comfort the emotional intensity of the user than two non-initiative responses. 
Furthermore, KEMI can generate more informative and specific responses with actual case knowledge.

\vspace{-0.05cm}

\section{Conclusions}
In this paper, we design a novel analysis framework for analyzing the feature of mixed initiative in ESC. The analysis demonstrates the necessity and importance of mixed-initiative interactions in ESC systems. To this end, we propose the KEMI framework to tackle the problem of mixed-initiative ESC. KEMI first retrieves actual case knowledge from a large-scale mental health knowledge graph with query expansion and subgraph retrieval. Then KEMI performs multi-task learning of strategy prediction and response generation with the retrieved knowledge. Extensive experiments show that KEMI outperforms existing methods on both automatic and human evaluation. The analysis also shows the effectiveness of incorporating actual case knowledge and the superiority of KEMI on the mixed-initiative interactions.

\section*{Limitations}\label{app:limit}
In this section, we analyze the limitations of this work:
\begin{itemize}[leftmargin=*]
    \item As it is the first attempt to analyze the mixed-initiative interactions in emotional support conversations, the proposed metrics can be further improved for more robust evaluation. 
    \item Since the knowledge retrieval is not the focus of this work, we did not spend much space on discussing the choice of different retrieval methods. As shown in Table~\ref{tab:ablation}, there is still much room for improving the knowledge retrieval from a large scale knowledge graph. It is also worth studying more efficient retrieval methods for retrieving knowledge from a densely connected KG. 
    \item The proposed method requires an additional mental health related knowledge graph constructed by experts or knowledgeable workers, which is probably difficult to obtain in some applications. However, different from other knowledge-intensive tasks that can be benefited from open-domain knowledge (\textit{e.g.,} Wikipedia), it attaches great importance in the professionals of the knowledge for building a helpful and safe ESC system.   
\end{itemize}

\section*{Ethical Considerations}\label{app:ethic}
The datasets adopted are publicly available and widely studied benchmarks collected from professionals or well-trained annotators. All personally identifiable and sensitive information, \textit{e.g.}, user and platform identifiers, in these dataset has been filtered out. We do not make any treatment recommendations or diagnostic claims. 
Compared with existing methods for emotional support conversations, the proposed method can be regarded as one step further to a more safer ESC system. 
The proposed method retrieves knowledge from a well-established mental health knowledge graph, which can be maintained by filtering out harmful information when applying into applications. Then the knowledge-enhanced approach can alleviate the randomness during the response generation and provide the guidance towards more positive responses. 
In order to prevent the happening of unsafe cases, the analysis of emotion intensity prediction can also serve as an alarming mechanism that calls for handoffs to an actual psychologist.

\bibliography{custom}
\bibliographystyle{acl_natbib}

\appendix
\section*{Appendix}

\section{Analysis of Mixed Initiative}\label{app:preliminary}

\subsection{Tools for Analysis}
We introduce two models that are adopted as off-the-shelf tools for the analysis of mixed initiative. 

\subsubsection{Utterance Initiative Classification}
To facilitate automatic analysis of utterance initiative, we train two utterance classification models by fine-tuning the pre-trained RoBERTa$_\text{large}$ model~\cite{roberta} on the ESConv~\cite{esconv} dataset, one for system utterance classification and the other for user utterance classification. 
We concatenate the previous utterance from another participant and the current utterance as the input for the binary initiative classification, either Initiative (I) or Non-Initiative (N). 
However, there is no initiative label in ESConv. Therefore, we manually annotate the initiative labels, I or N, for each utterance according to the EAFR schema. 
The resulting dataset contains $\sim$38K utterance-label pairs (E: 13K, A: 9K, F: 7K, R: 10K).

\subsubsection{Emotion Intensity Prediction}\label{app:emo_int}
Similarly, we also fine-tune an emotion intensity prediction model $f_e(\cdot)$ based on the pre-trained RoBERTa$_\text{large}$ model~\cite{roberta} on the ESConv~\cite{esconv} dataset. 
Given a user utterance, the model aims to predict the negative emotion intensity level $e_{ij} = f_e(u_{ij})$, ranging from 1 to 5, which indicates the user's emotional state. 
In ESConv, the initial and final emotion intensity levels of the user have already been annotated. 
Therefore, we regard the first user utterance after greetings to match with the initial emotion intensity, while the last user utterance before goodbyes to match with the final emotion intensity. 
The resulting dataset contains 2,450 utterance-label pairs (1: 331, 2: 506, 3: 557, 4: 629, 5: 427). 

\subsubsection{User Simulator}\label{app:usersim}
Inspired by the evaluation of mixed-initiative CIS~\cite{wsdm22-mix-usersim}, we simulate a user based on a large-scale generative language model, namely BlenderBot~\cite{blenderbot}. 
In our case, we fine-tune a semantically-conditioned generation model $g(\cdot)$, guided by the underlying problematic situation:
\begin{equation}\small
    p_g(\bm{a}|s,\mathcal{C},r) = \prod\nolimits^L_{l=1} p_g(a_l|a_{<l};s,\mathcal{C},r),
\end{equation}
where $a$ is the user's feedback to the generated response $r$.  
The generation model is fine-tuned on the whole dataset, including the test set. If the ESC system generates a perfect response, the user simulator should give the ground-truth feedback as the real user. 

We adopt the same utterance initiative classification model and emotion intensity prediction model $f_e(\cdot)$ described in Appendix~\ref{app:emo_int} to annotate the generated response. 
The annotation results are used for calculating the emotional support metrics. 
In particular, the calculation of Relaxation metric involves the user's emotion intensity after receiving the generated response. 
The user simulator $g(\cdot)$ is employed to simulate the user's feedback. 
Then the calculation of the Relaxation metric in Eq.(\ref{eq:rel}) becomes:
\begin{equation}\small
    \text{Rel}_{i}[r_{i}=S] = f_e(u_{<i}[r_{<i}=U]) - f_e(g(s,\mathcal{C},r)). \label{eq:sim_rel}
\end{equation}

\begin{figure}
\centering
\includegraphics[width=0.48\textwidth]{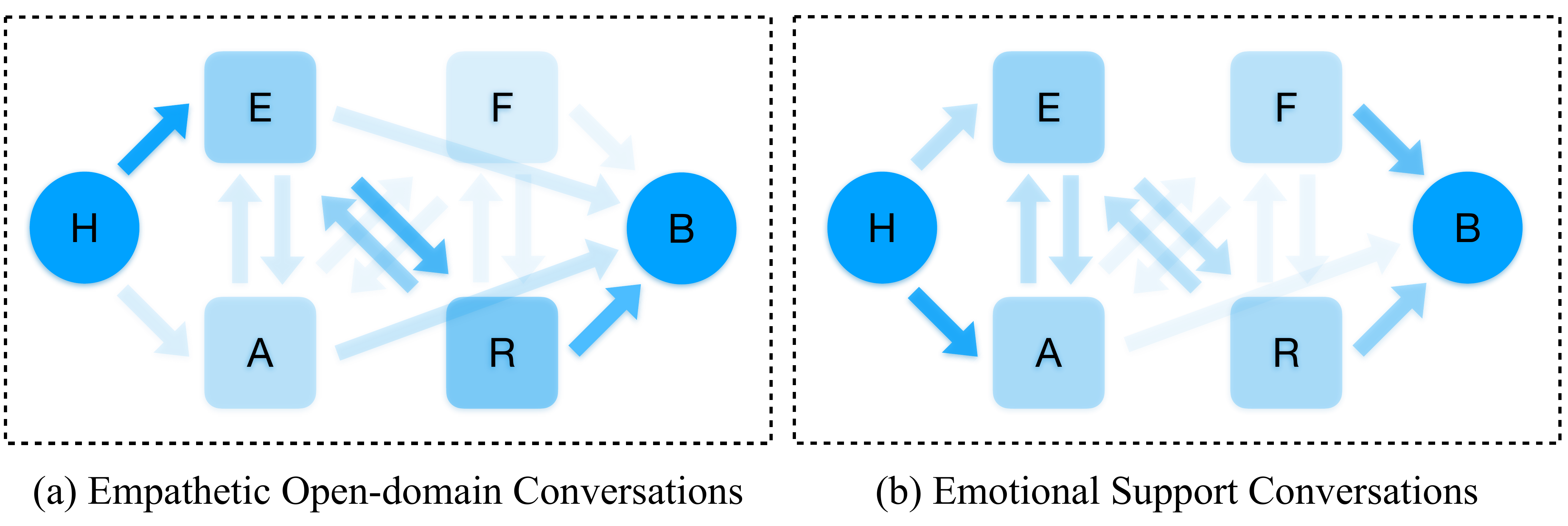}
\caption{Dialogue flow. H, B, E, A, F, and R denote Hi, Bye, Expression, Action, Feedback, and Reflection. The color intensity denotes the proportions of the utterance labels and the initiative transitions.}
\label{dialog_flow}
\end{figure}
\subsection{Dialogue Flow}\label{app:dial_flow}
Following previous studies on mixed-initiative CIS systems~\cite{qrfa,tois21-mix}, we draw the dialogue flow diagram to observe the initiative switch patterns between dialogue participants in ED and ESC. 
As shown in Figure~\ref{dialog_flow}, the circles represent the beginning and ending of the dialogue, while the boxes represent the EAFR utterance labels. The color intensity denotes the proportions of the utterance labels and the initiative transitions. 
There are several notable observations: 
(i) As for the proportion of EAFR labels, \textit{Expression} and \textit{Reflection} constitute the majority of the utterances in ED, while four labels are more equally distributed in ESC. 
(ii) As for the beginning and ending of dialogues, users are more often to take the initiative to start a conversation in ED, and the dialogue will be ended by the system. Differently, in ESC, the conversation is usually started by the system. 
(iii) As for the initiative switches, most of cases in ED are that users express their feelings and then the system tries to comfort them with empathy. However, the proportion of each type of initiative transitions in ESC is relatively equal. 
Therefore, we conclude that \textbf{the system in ED generally serves as a passive role, while the system in ESC needs to switch the initiative role during the conversation}.

\subsection{Conversation Progress}\label{app:conv_prog}
\begin{figure}
     \centering
     \includegraphics[width=0.48\textwidth]{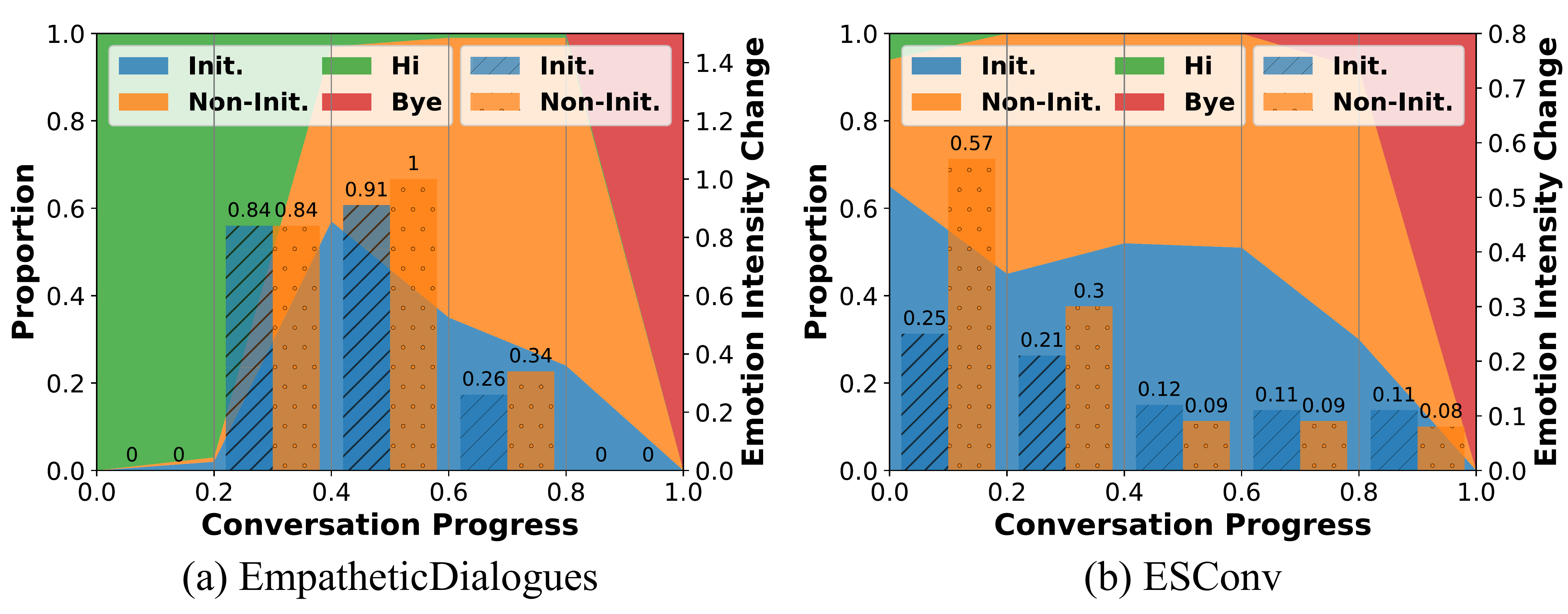}
     \caption{The distribution of utterance initiative (the stack plot) and the emotion intensity change (the bar chart) at different conversation progress.}
     \label{progress}
\end{figure}

We analyze the conversation progress by dividing the whole conversation into five equal length intervals. 
To alleviate the noise from greeting (Hi) and farewell (Bye) utterances, we heuristically identify these utterances by rules, \textit{e.g.}, containing ``Hi/Hello'' at the beginning or ``Bye/Goodbye'' at the end of the conversation. 
Specifically, we compute the distribution of initiative labels for system utterances and the average change of emotion intensity levels at each conversation phase. 
As shown in Figure~\ref{progress}, under both cases, the system tends to take the initiative at the beginning of the conversation for exploring the user's problem, while acting passively at the latter stage of the conversation. 
Interestingly, at the early phase of conversations, compared with non-initiative utterances, system-initiative ones fail to relax the emotion intensity of help-seekers in ESC. 
This is because the request for information from users to understand their problems  is likely to raise users' negative emotions.  
However, at the latter stage of the conversation, initiative utterances can better lower down users' intensity levels, leading to a higher emotional intensity change rate than non-initiative ones. 
This indicates that \textbf{(i) the timing for system-initiative interactions is important, and (ii) it is more helpful to provide suggestions or information for users to solve the problem when the emotion of users has been eased}.

\begin{table}
    \centering
    \begin{adjustbox}{max width=0.48\textwidth}
    \setlength{\tabcolsep}{0.5mm}{
    \begin{tabular}{l|cc|ccc|ccc|ccc}
    \toprule
    \multicolumn{1}{c}{}     
    & \multicolumn{2}{c}{Proactivity} &\multicolumn{3}{c}{Information}&\multicolumn{3}{c}{Repetition} &\multicolumn{3}{c}{Relaxation} \\
    \cmidrule(lr){2-3}\cmidrule(lr){4-6}\cmidrule(lr){7-9}\cmidrule(lr){10-12}
    
    \multicolumn{1}{c}{} & Init. & \multicolumn{1}{c}{Non.}  & Init. & Non. & \multicolumn{1}{c}{All}& Init. & Non. & \multicolumn{1}{c}{All}& Init. & Non. & \multicolumn{1}{c}{All} \\
    \midrule
    ED & 0.28 & 0.72& 2.14 & 2.69 & 2.46& 0.42 & 0.44 & 0.43& 0.83 & 0.82 & 0.83\\ 
    ESC & 0.48 & 0.52 & 3.32 & 3.06 & 3.19& 1.06 & 1.18 & 1.12& 0.16 & 0.20 & 0.18\\
    \bottomrule
    \end{tabular}}
    \end{adjustbox}
    \caption{Comparisons on emotional support metrics.}
    \label{tab:metrics}
\end{table}

\subsection{Emotional Support Metrics} \label{app:metrics}
Table~\ref{tab:metrics} summarizes the scores of the emotional support metrics introduced in Section~\ref{sec:eafr} for two datasets. 
Firstly, the proportion of system-initiative interactions in ESC is much higher than that in ED, showing the importance of mixed initiative in ESC systems. 
Secondly, system-initiative utterances provide more information than non-initiative utterances in ESC, while an opposite result is observed in ED. This shows that the ESC system provides informative responses when taking the initiative. 
Thirdly, in both datasets, there is more repetitive information in non-initiative system utterances than initiative ones, indicating that reflection is more important in non-initiative interactions. 
Last but not least, the average \textit{relaxation} score in ESC is much lower than that in ED. 
We attribute this to two reasons: (i) Empathetic responses have more positive effects on the user's emotions. (ii) The system-initiative interactions sometimes may increase the user's emotion intensity, as discussed in Appendix~\ref{app:conv_prog}.

\section{Definition of COMET Relations}\label{app:comet}
We adopt five types of commonsense relations in \texttt{COMET}~\cite{comet}, whose original definitions are as follows: 
\begin{itemize}[leftmargin=*]
    \item \texttt{xEffect}: The effect that the event would have on Person X.
    \item \texttt{xIntent}: The reason why X would cause the event.
    \item \texttt{xNeed}: What Person X might need to do before the event.
    \item \texttt{xReact}: The reaction that Person X would have to the event.
    \item \texttt{xWant}: What Person X may want to do after the event.
\end{itemize}

\section{Details of HEAL}\label{app:heal}
\begin{table}
    \centering
    \begin{adjustbox}{max width=0.48\textwidth}
    \setlength{\tabcolsep}{1.5mm}{
    \begin{tabular}{ccccc}
    \toprule
    & Stressor & Expectation & Response & Affect. State \\
    \midrule
    Stressor & \textbf{4,363} & 9,801 & - & - \\
    Expectation & 9,801 & \textbf{3,050} & 26,628 & 3,050 \\
    Response & -& 26,628 & \textbf{13,416} & -\\
    Affect. State &- & 3,050 & -& \textbf{41} \\
    \bottomrule
    \end{tabular}}
    \end{adjustbox}
    \caption{Statistics of \texttt{HEAL} adopted in our experiments.}
    \label{tab:heal}
\vspace{-0.3cm}
\end{table}

\texttt{HEAL}~\cite{heal} is a knowledge graph developed upon 1M distress discussions and their corresponding consoling responses curated from mental health support conversations. 
It consists of 22K nodes with five different types: \textit{stressors}, \textit{expectations}, \textit{responses}, \textit{feedback}, and \textit{affective states} associated with distress dialogues, and forms 104K connections between different types of nodes. 
The statistics of the adopted \texttt{HEAL} are presented in Table~\ref{tab:heal}.

\section{Baselines}\label{app:baseline}
We provide extensive comparisons with the following strong baselines, including both non-PLM and PLM-based methods: 
\begin{itemize}[leftmargin=*]
    \item Transformer~\cite{transformer} for Seq2Seq response generation.   
    \item MoEL~\cite{emnlp19-moel} is a Transformer-based model that involves multi-decoders to enhance the empathy for different emotions. 
    \item MIME~\cite{emnlp20-mime} is a Transformer-based model that mimics the emotion of the speaker for empathetic response generation. 
    \item BlenderBot~\cite{blenderbot} is an open-domain dialogue model pretrained with multiple skills, including empathetic responding. BlenderBot-Joint~\cite{esconv} jointly predicts strategies and generates responses. 
    \item GLHG~\cite{ijcai22-glhg} is a BlenderBot-based model, which employs a hierarchical graph network to encode multi-source information.
    \item MISC~\cite{acl22-misc} is a BlenderBot-based model, which incorporates commonsense knowledge and mixed support strategy to jointly predicts support strategies and generates responses.  
\end{itemize}

\end{document}